\documentclass[letterpaper, 10 pt, conference]{ieeeconf}  
                                                   
\usepackage{color}
\usepackage{hyperref}
\usepackage{booktabs}
\usepackage{indentfirst}
\usepackage{multirow}
\usepackage{graphicx}
\usepackage{amsmath}
\usepackage{amssymb}
\usepackage{makecell}

\IEEEoverridecommandlockouts                              
                                                          
\title{MBPU: A Plug-and-Play State Space Model for Point Cloud Upsamping with Fast Point Rendering
}

\author{Jiayi Song, Weidong Yang, Zhijun Li, \IEEEmembership{Fellow, IEEE}, Wen-Ming Chen, Ben Fei
\thanks{Jiayi Song and Weidong Yang are with the School of Computer Science, Fudan University, Shanghai, China, 200433 (e-mails:  22307130359@m.fudan.edu.cn; wdyang@fudan.edu.cn)}%
\thanks{Ben Fei is with the Department of Information Engineering, The Chinese University of Hong Kong, Hong Kong, China (e-mail: benfei@cuhk.edu.hk)}%
\thanks{Wen-Ming Chen is with the Academy for Engineering and Technology, Fudan University, Shanghai 200433, China (e-mail: chenwm@fudan.edu.cn).}
\thanks{Zhijun Li is with the School of Mechanical Engineering, Tongji University, Shanghai, 200092, China (e-mail: zjli@ieee.org).}
}

\begin{document}

\maketitle
\thispagestyle{empty}
\pagestyle{empty}

\begin{abstract}

The task of point cloud upsampling (PCU) is to generate dense and uniform point clouds from sparse input captured by 3D sensors like LiDAR, holding potential applications in real yet is still a challenging task. 
Existing deep learning-based methods have shown significant achievements in this field. However, they still face limitations in effectively handling long sequences and addressing the issue of shrinkage artifacts around the surface of the point cloud.
Inspired by the newly proposed Mamba, in this paper, we introduce a network named MBPU built on top of the Mamba architecture, which performs well in long sequence modeling, especially for large-scale point cloud upsampling, and achieves fast convergence speed.
Moreover, MBPU is an arbitrary-scale upsampling framework as the predictor of point distance in the point refinement phase.
At the same time, we simultaneously predict the 3D position shift and 1D point-to-point distance as regression quantities to constrain the global features while ensuring the accuracy of local details.
We also introduce a fast differentiable renderer to further enhance the fidelity of the upsampled point cloud and reduce artifacts. 
It is noted that, by the merits of our fast point rendering, MBPU yields high-quality upsampled point clouds by effectively eliminating surface noise.
Extensive experiments have demonstrated that our MBPU outperforms other off-the-shelf methods in terms of point cloud upsampling, especially for large-scale point clouds.

\end{abstract}


\section{Introduction}

\label{sec:introduction}


%
%
%
%
Point cloud, consisting of a set of points in a three-dimensional coordinate system, is a fundamental representation of real-world objects captured through various sensing technologies like LiDAR scanners, depth cameras, or photogrammetry~\cite{leberl2010point,fei2022comprehensive,fei20243d}. 
They have wide-ranging applications in fields such as robotics, autonomous vehicles, augmented reality, urban planning, and environmental modeling~\cite{rozenberszki2020lol,guo2020deep}.
While point clouds provide valuable insights into the geometry and structure of objects or scenes, they often suffer from irregular sampling densities. 
And the irregularity in point clouds often poses challenges for downstream tasks such as surface reconstruction~\cite{bernardini1999ball,fei2023dctr}, object recognition~\cite{bai2023rangeperception,nguyen2024language}, and semantic segmentation~\cite{che2019object,schult2023mask3d}.
This is mainly due to existing algorithms may not perform uniformly well across different densities of point clouds.
\par To address the challenges posed by irregular sampling densities in point clouds, point cloud upsampling (PCU) serves as a vital preprocessing step~\cite{rong2024repkpu,li2024learning,qu2024conditional}. 
PCU involves generating additional points within the point cloud, enriching the uniformity and density of point clouds while preserving their inherent structure and features~\cite{jung2022fast}.
By increasing the density of points, PCU aims to reduce sampling artifacts, enhance geometric details, and facilitate a more accurate and robust analysis of the underlying geometry, thereby improving the quality and reliability of subsequent processing and analysis~\cite{bai2022bims}. 

\par In recent years, significant research endeavors~\cite{li2019pu,qian2020pugeo,yifan2019patch,li2021point,ye2021meta,zhao2021sspu,mao2022pu,li2022semantic} have been dedicated to advancing PCU techniques.
Deep learning-based approaches, in particular, have shown promising results by learning complex patterns and relationships from large-scale point cloud datasets~\cite{luo2021pu,qian2021pu,qiu2022pu,feng2022neural,he2023grad}. 
For example, PU-Net~\cite{yu2018pu} utilizes PointNet++ as a backbone to learn multi-level features and expand them via multi-branch MLPs.
MPU~\cite{yifan2019patch} employs multiscale information to generate points from local to global scales progressively.
PU-GAN~\cite{li2019pu} adopts a generative adversarial network framework to learn to generate points with a uniform distribution.
Dis-PU~\cite{li2021point} introduces a dense generator and spatial refiner to deal with the upsampling process.
PU-GCN's NodeShuffle technique uses a Graph Convolutional Network (GCN) to enhance the encoding of local point information~\cite{qian2021pu}.
Additionally, Grad-PU~\cite{he2023grad} integrates P3DConv and a gradient descent process with learned distance functions to achieve arbitrary-scale point cloud upsampling.
\par While existing PCU methods have made significant progress in the quality of point cloud upsampling, there are still two main challenges to be addressed. 
First, existing PCU methods still struggle to handle long sequences. 
When the scale of point clouds grows, the computational complexity and memory consumption of these PCU methods increase significantly.
Moreover, some methods will experience a significant decrease in performance and upsampling quality while processing long sequence data due to network capacites.
Second, due to insufficient understanding and lack of constraints on surface structures during the upsampling process, existing models tend to produce noise points on the surface of point clouds, resulting in shrinkage artifacts~\cite{luo2021score}.

\par In this paper, to tackle these challenges, we are inspired from Mamba~\cite{gu2023mamba,zhou20243dmambaipf} in language models and propose a plug-and-play state space model for point cloud upsampling with fast point rendering, dubbed MBPU.
Taking low-resolution point clouds as input, our model applies midpoint interpolation and employs a dense network that combines MLPs and Mamba module~\cite{liang2024pointmamba} to extract local and global features.
The network also estimates the discrepancy between interpolated point clouds and ground truth with Mamba as the kernel, which leverages the linear complexity of SSM, resulting in faster convergence speed and better performance while dealing with long sequences.
The discrepancy between interpolated point clouds and ground truth is estimated through two branches -- 1D point-wise distance prediction and 3D position shift -- to better restore local features while preserving global structure.   
Also, we introduce a fast differentiable rendering module~\cite{zhang2021progressive} to the 3D position shift branch, which helps us more effectively reduce outliers and ensure the fidelity of the point cloud. 
Guided by predictions from two branches, MBPU then progressively performs distance regression to refine and achieve the final upsampled point cloud.
Experimental results demonstrate that our MBPU is able to upsample point clouds with better performance in handling long sequences, faster convergence, and fewer shrinkage artifacts. 
\par Our main contributions can be summarized as:
\begin{itemize}
\item[$\bullet$] Our MBPU proposes a plug-and-play state space model for point cloud upsampling for the first time, achieving better performance on long sequences and faster convergence and training speed.
\item[$\bullet$] We devise a two-branches prediction mechanism for 1D point-wise distance prediction and 3D position shift, which not only ensures the fidelity of local details but also maintains global structure.
\item[$\bullet$] We also introduce a fast and effective point-based differentiable renderer, which reduces outliers and artifacts by the merits of the supervisory from rendered 2D images between upsampled point cloud with 3D position shift and ground truth.
\end{itemize}

\section{Methods}
\label{sec:methods}

\subsection{Overview}

\begin{figure}[t]
    \centering
    \includegraphics[width=\linewidth]{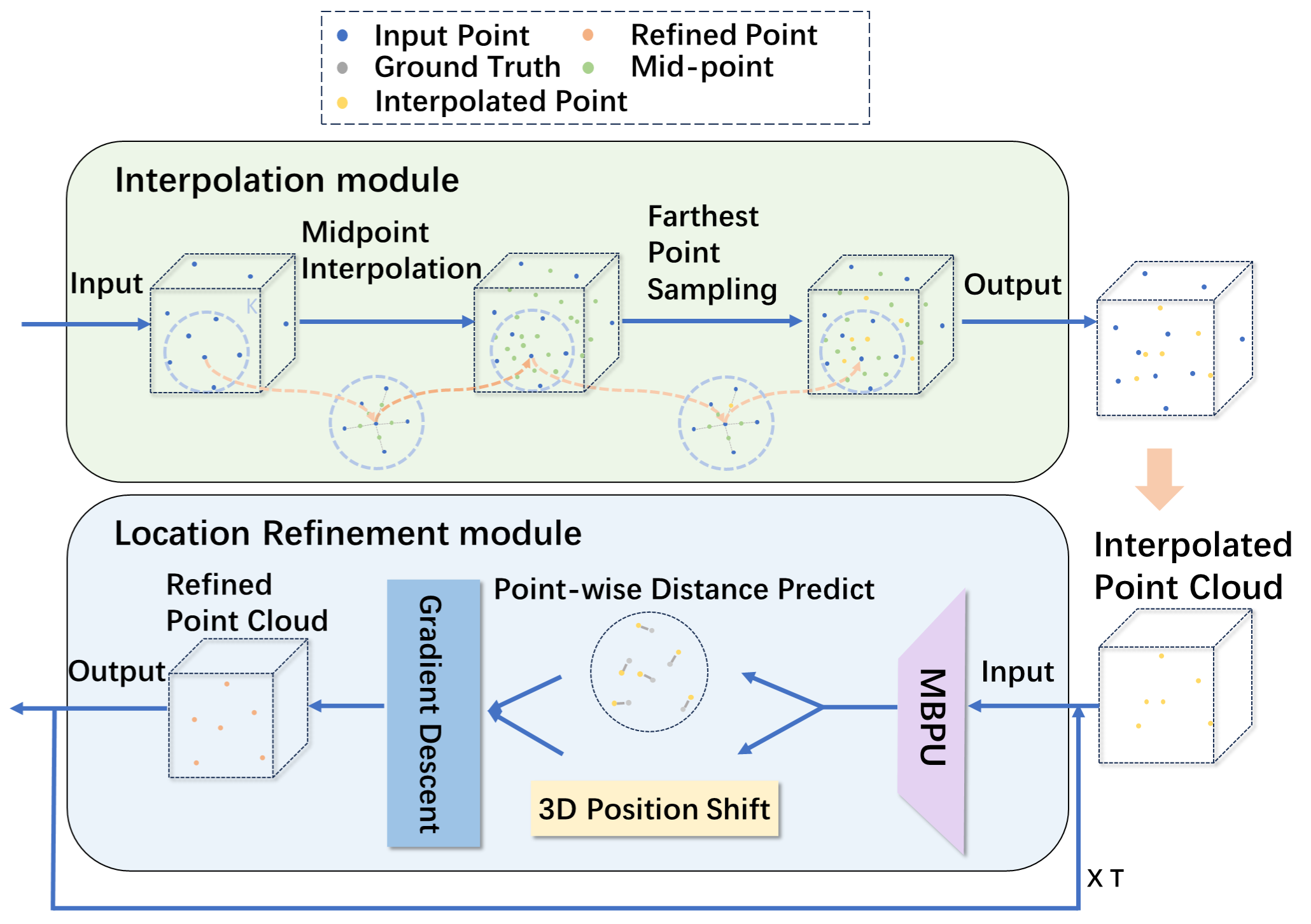}
    \vspace{-0.9cm}
    \caption{Pipeline of our framework. Given a low-resolution input point cloud, we first perform midpoint interpolation and FPS on it to obtain point clouds with the desired number of points. Then the interpolated point cloud is fed into our MBPU network, for gradient descent to progressively refine the interpolated point cloud.}
    \label{fig:pipeline}
\vspace{-0.5cm}
\end{figure}

Inspired by Mamba~\cite{liang2024pointmamba,li2024videomamba}, we propose MBPU to upsample point clouds with arbitrary ratios, which demonstrates great performance on long sequences, avoids shrinkage artifacts around the surface and achieves fast convergence speed. 
The pipeline of our MBPU is shown in Fig.\ref{fig:pipeline}. 
It contains a midpoint interpolation module, which generates an interpolated point cloud containing an arbitrary number of interpolation points, and a location refinement module which refines the location of points.

\par Specifically, we first perform midpoint interpolation on the input low-resolution point cloud to obtain an interpolated point cloud with our desired upsampling rate. 
Then we refine the positions of the points in the interpolated point cloud using gradient descent, allowing them to progressively converge towards the ground truth.
During this process, our MBPU is trained to infer 3D position shift and 1D point-wise distance between the interpolated point cloud and ground truth, thereby guiding the refinement process.
Besides, a differentiable rendering module is adopted in the 3D position shift prediction process to decrease outliers and artifacts.
Then, the point cloud is iteratively refined through gradient descent to obtain the final unsampled point cloud.
\subsection{Midpoint Interpolation}

We employ the midpoint interpolation method to upsample the initial point cloud. 
This method allows us to generate an interpolated point cloud with any desired number of points, thereby enabling our model to achieve its goal of implementing a flexible upsampling rate. 
In this module, we start by taking a low-resolution point cloud $\boldsymbol{P_{lres}}$ as input. For each point $\boldsymbol{\mathit{p}\in P_{lres}}$ in the input point cloud, its k-nearest points $\boldsymbol{\mathit{p_{i}(i=1,2,...,k)}}$ are first identified.
These k points are then used for midpoint interpolation with the corresponding point $\boldsymbol{\mathit{p}}$ in the input point cloud, resulting in a set of interpolated points. 
\begin{equation}\small
\label{eq1}
    \boldsymbol{\mathit{p_{mid_i}=(p+p_{i})/2 \qquad(i=1,2,...,k)}}.
\end{equation}

\par Subsequently, to eliminate potential duplicate points and control the number of points based on the desired sampling rate, this module performs furthest point sampling on the interpolated points.
Specifically, it calculates offsets between input points and interpolated points.
Subsequently, farthest point sampling is conducted to determine the sampled point indices, based on the desired number of points to sample. 
These indices are then used to extract the corresponding sampled points from the interpolated point cloud.
Thus, the combination of midpoint interpolation and furthest point sampling forms the final interpolated point cloud $\boldsymbol{\mathit P_{I}}$ with the desired upsampled number of points.

\begin{figure}[t]
    \centering
    \includegraphics[width=\linewidth]{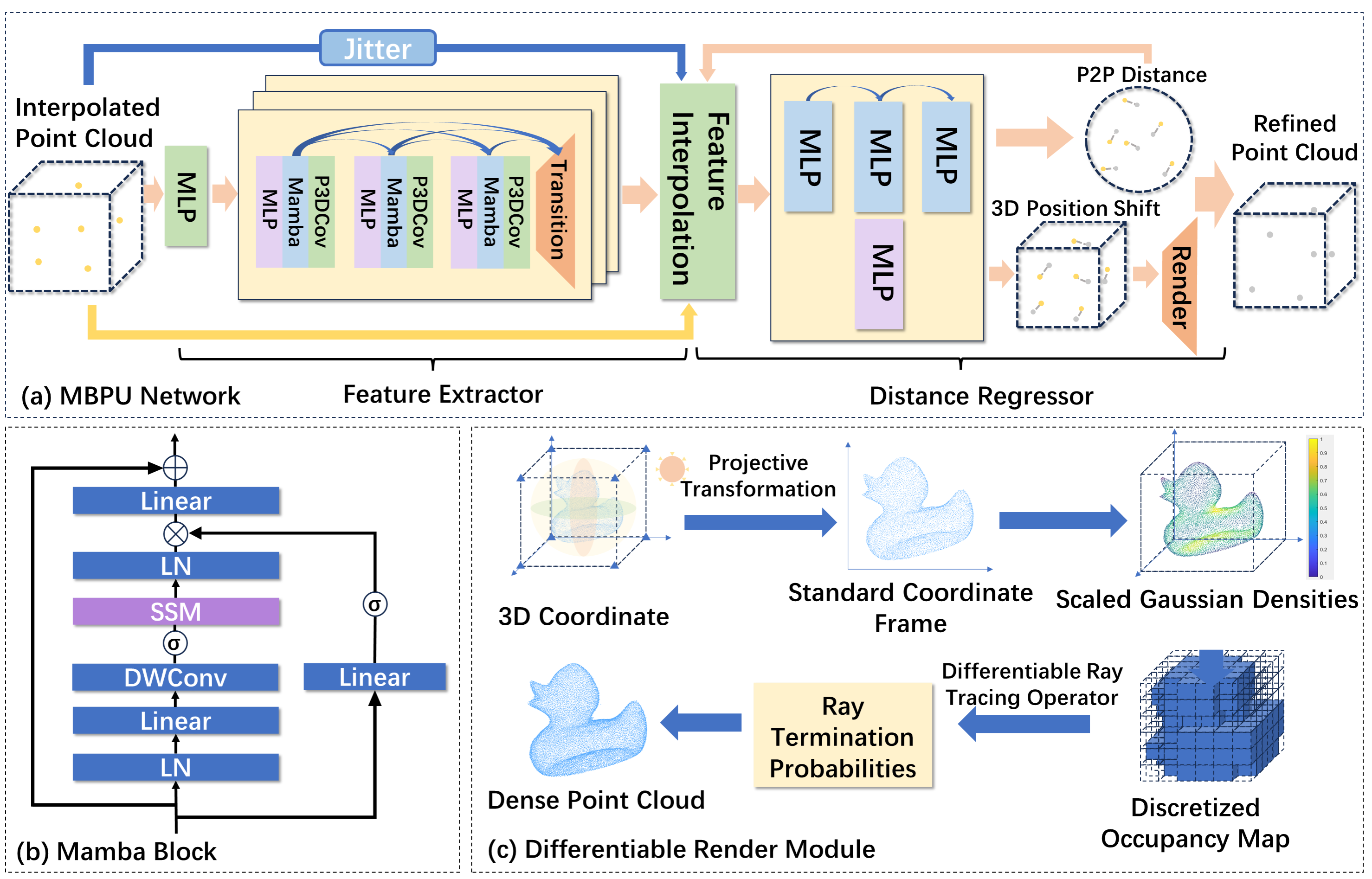}
    \vspace{-0.8cm}
    \caption{(a) The overall architecture of our MBPU, mainly consists of two modules: a feature extractor and a distance regressor. In the feature extractor, We utilize three mixer modules in each dense block to extract local features and a transition layer to reduce channel.
    In the distance regressor, we estimate 3D position shift and  P2P distance through two branches. 
    (b) Structure of Mamba block~\cite{liang2024pointmamba}, which consists of layer normalization (LN), Selective SSM, depth-wise convolution (DW), and MLPs. 
    (c) Pipeline of our devised differentiable render module, which renders depth images of temporary upsampled point clouds and ground truth. The view loss between these rendered images will be back-propagated to update the parameters of the network.
    The small triangles in the first cube represent the camera poses.}
    \label{fig:network}
\vspace{-0.5cm}
\end{figure}

\begin{figure*}[t]
    \centering
    \includegraphics[width=\linewidth]{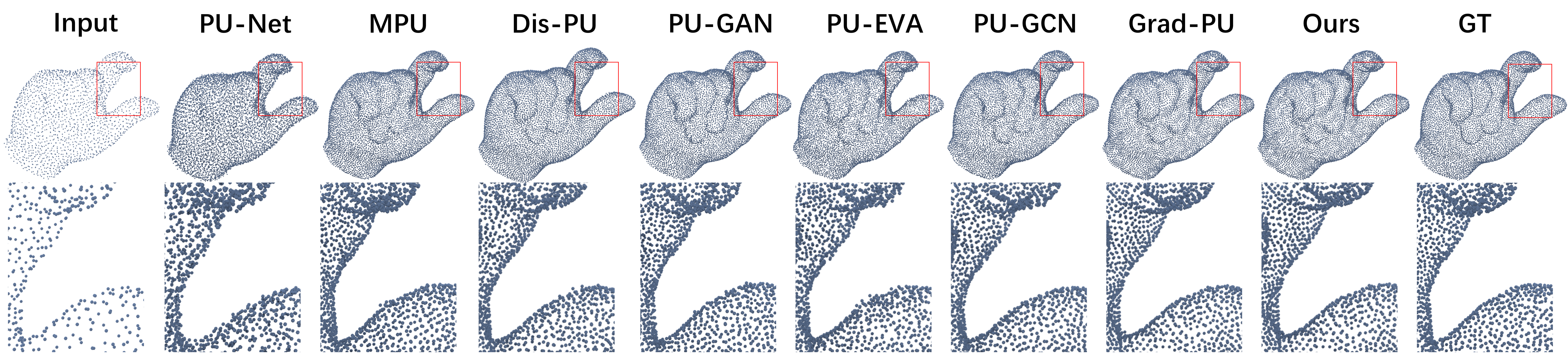}
    \vspace{-0.8cm}
    \caption{$4\times$ visualization results on PU-GAN dataset. Our method performs less outliers and more fine-grained details. }
    \label{fig:4X-PUGAN}
\vspace{-0.3cm}
\end{figure*}

\begin{figure*}[t]
    \centering
    \includegraphics[width=0.9\linewidth]{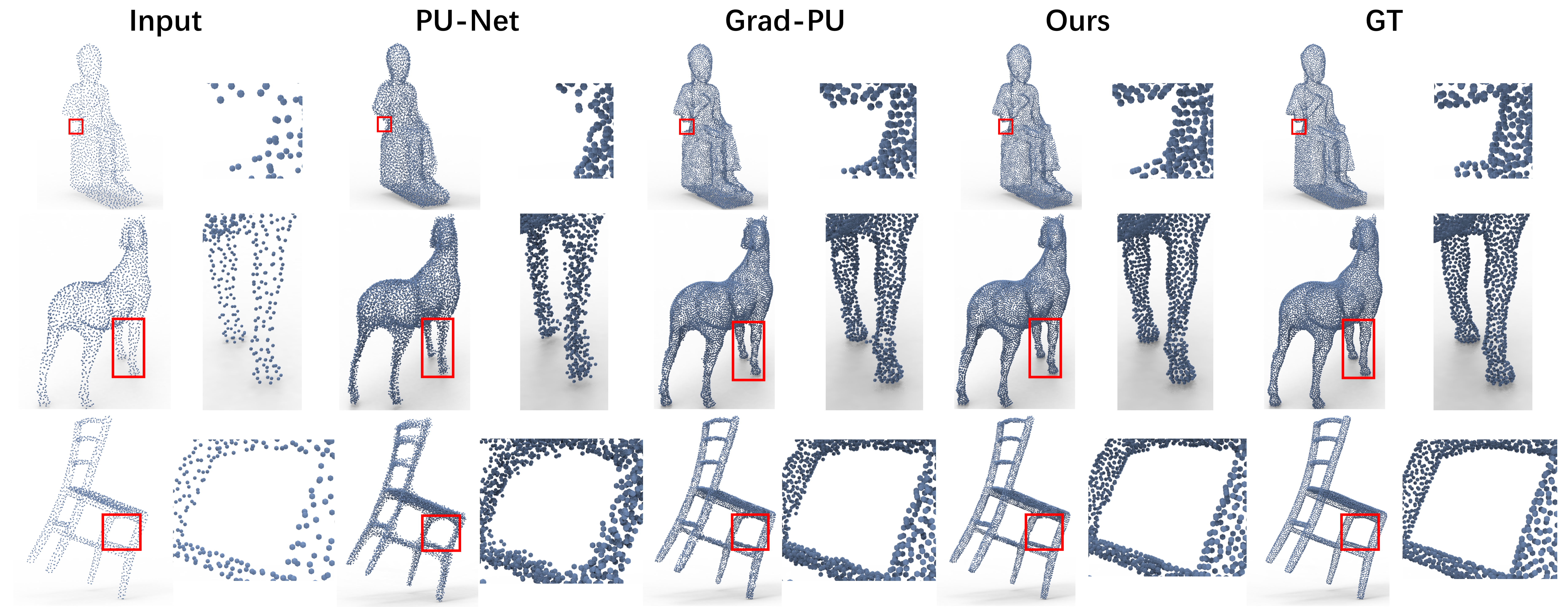}
    \vspace{-0.4cm}
    \caption{Visualization results on PU1K dataset with $4\times$ upsampling rate. Our method demonstrates more accurate details and global structure. }
    \label{fig:4X-PU1K}
\vspace{-0.3cm}
\end{figure*}

\subsection{Location Refinement Module}
After obtaining interpolated point cloud $\boldsymbol{\mathit P_{I}}$ from the midpoint interpolation module, we use the point location refinement module to refine the location of the points, with the objective of minimizing the differences between interpolated point cloud and ground truth. 
\par We utilize the gradient descent to refine the positions of points, whose formulation is as below:
\begin{equation}\small
\label{eq2}
\boldsymbol{\mathit{\nabla{F(p)}=\Big(\frac{\partial F}{\partial p^{i}}(p)\Big) }}, 
\end{equation}
where $\boldsymbol{\mathit{i}}$ represents the i-th dimension of the varibale $\boldsymbol{\mathit{p}}$,
\begin{equation}\small
\label{eq3}
\boldsymbol{\mathit{p_{t+1}=p_{t}-\lambda\nabla{F(p_{t})}\qquad{(t=0,1,...,T-1)} }}.
\end{equation}

Given the initial interpolated point $\boldsymbol{\mathit p_{0}\in P_{I}}$, in this module, this point will be iterated $\boldsymbol{\mathit T}$ times, with each iteration progressively approaching the ground truth point by following the gradient direction of the distance function  $\boldsymbol{\mathit F(p)}$ between the interpolated point and the corresponding nearest ground truth point. 
During this process, the distance function between the interpolated point and the ground truth point is determined jointly by the one-dimensional displacement between them and the disparity in the 3D position shift. 
However, since our model is not allowed to directly access the ground truth point cloud during testing, the distance function here is effectively approximated by the differentiable approximation of a pre-trained network.

\subsection{MBPU for Distance Function Learning}

We have developed an MBPU network specifically designed to estimate the aforementioned distance function, which mainly includes two modules: Mamba-based Feature Extractor and Distance Regressor.
By leveraging Mamba as the kernel, the MBPU enables seamless integration and practical utilization of point cloud upsampling.
\subsubsection{Mamba-based Feature Extractor}
\par With the interpolated point cloud gained from the interpolation module, we firstly obtain the initial features of the point cloud through an MLP.
Then, the features flow through three dense blocks, which consist of three densely connected mixer modules and a transition layer.
The mixer module contains an MLP to reduce the dimensionality of features, a Mamba block, and P3DConv~\cite{he2023grad} to extract and integrate features.
\par The new features obtained in the mixer module are then passed through the transition layer, reducing the size and number of channels in the feature map, and thereby achieves compression and filtering of information.
\par Then, the local features $\left\{ \boldsymbol{\mathit{l_{0},l_{1},l_{2},l_{3}}}\right\}$ obtained through multiple dense blocks undergo max pooling to obtain global features $\boldsymbol{\mathit{g}}$.
Finally, local and global features, together with the initial interpolated point cloud, are used to conducted feature interpolation.

\subsubsection{Distance Regressor}
For each point in the point cloud, the distance function is estimated on the local features and global features. 
After gaining the local and global features in the Mamba-based Feature Extractor, we interpolate features with information from the initial interpolated point cloud, local and global features, and 1D point-to-point distance fed back, and then cat the features together.
Then we conduct two branches on the interpolated features to progressively refine the point cloud. 
\par The first branch is based on a 1D point-wise predictor, in which, a point-to-point distance is estimated with a four-layer MLP. 
Given the concatenated feature $\boldsymbol{\mathit{X\in R^{b\times c\times n}}}$, the output $\boldsymbol{\mathit{Output\in R^{b\times 1\times n}}}$ is as the equation below:
\begin{equation}\small
\label{eq4}
\boldsymbol{\mathit{Output= MLP_3(MLP_2(MLP_1(MLP_0(X)))) }},
\end{equation}
where $\boldsymbol{\mathit{b}}$ represents batch size, $\boldsymbol{\mathit{c}}$ represents the number of feature channels, $\boldsymbol{\mathit{n}}$ represents the dimensionality of features and $\boldsymbol{\mathit{MLP_{i}}}$ denotes the $\boldsymbol{\mathit{i}}$-th multi-layer perceptron module and its activation function.
\par Simultaneously, to guarantee the accuracy of global features and maintain structure, we introduced the second branch to further refine the positions of the points.
Still with four MLPs but the last one restoring features to three dimensions, the 3D position shift predictor in the second branch processes the interpolated features obtained from the third MLP in the first branch through the last different MLP to gain the 3D position shift.
\par The formulation of the second branch is as below, with the same concatenated feature  $\boldsymbol{\mathit{X\in R^{b\times c\times n}}}$ as input and a different $\boldsymbol{\mathit{Output’\in R^{b\times 3\times n}}}$ as output:
\begin{equation}\small
\label{eq5}
\boldsymbol{\mathit{Output’= MLP_3’(MLP_2(MLP_1(MLP_0(X)))) }}.
\end{equation}

\par The prediction results from the 3D position shift branch are fed to a fast differential rendering module in Section~\ref{subsec:render} to decrease outliers and artifacts.
Finally, the point-to-point distance acts together with the 3D position shift prediction to control the refinement process of the point cloud. 
\subsubsection{Mamba}
\par Our Mamba block~\cite{liang2024pointmamba} takes advantage of SSM to help our MBPU better handle long sequences and achieve fast convergence speed. As Fig.~\ref{fig:network}(b) shows, its structure consists of layer normalization (LN), Selective SSM, depth-wise convolution (DW), and residual connections. 
\par Specifically, the output features $\boldsymbol{\mathit{F}}$ from MLP in the mixer module are fed into a network containing one layer of Mamba block. The structure of a Mamba block is expressed in the following formula:
\begin{equation}\small
\label{eq8}
\boldsymbol{\mathit{T=SSM(\sigma(DW(MLP(LN(F))))) }},
\end{equation}
\begin{equation}\small
\label{eq9}
\boldsymbol{\mathit{S=MLP(LN(T)\times\sigma(LN(F))+F }},
\end{equation}
where $\boldsymbol{\mathit{S}}$ represents the output of Mamba block, $\boldsymbol{\mathit{\sigma}}$ indicates SiLU activation.

\subsection{Point Rendering}
\label{subsec:render}

In order to enhance the accuracy of the regression for 3D position shifts, we have developed a fast and differentiable point rendering module.
The 3D position shift is applied to the interpolated point cloud to generate a temporary upsampled point cloud. 
Then, this temporary upsampled point cloud and ground truth go through our devised point rendering to obtain depth images, which contain rich information regarding the overall 3D point clouds and shape edges.
The loss gradient calculated between these two sets of depth images will be used to update the network parameters. This process effectively helps to eliminate shrinkage artifacts and noise surrounding the upsampled point clouds.

The pipeline of point cloud render is shown in Fig.~\ref{fig:network}(c). 
The render first projects the 3D coordinates of points in one of the 32 camera positions to a standard frame. 
With all of the 32 standard frames, scaled Gaussian densities are obtained to describe the discretized points, which are then formed in the occupancy map.
Then the discretized occupancy map is transformed into ray termination probabilities, utilizing a differentiable ray tracing operator.
Finally, a set of dense point cloud images is obtained by projection.
\par Considering the ground truth view images $\boldsymbol{\mathit{GT}}$ and reconstructed view images $\boldsymbol{\mathit{T}}$ of the $i$-th camera positions being projected as $\boldsymbol{\mathit{R_{GT}^{i}}}$ and $\boldsymbol{\mathit{R_{T}^{i}}}$, the rendering loss $\boldsymbol{\mathit{L_{v}}}$ can be described as follows:
\begin{equation}
\label{eq10}
\boldsymbol{\mathit{L_{v}=\sum_{i=1}^{I}\sum_{x=1}^{W}\sum_{y=1}^{H}\left|R_{GT}^{i}(x,y)-R_{T}^{i}(x,y)\right| }},
\end{equation}
where $\boldsymbol{\mathit{I}}$ represents the total number of camera positions, $\boldsymbol{\mathit{W}}$ and $\boldsymbol{\mathit{H}}$ represent the width and height of the projected image respectively.


\begin{figure*}[t]
    \centering
    \includegraphics[width=\linewidth]{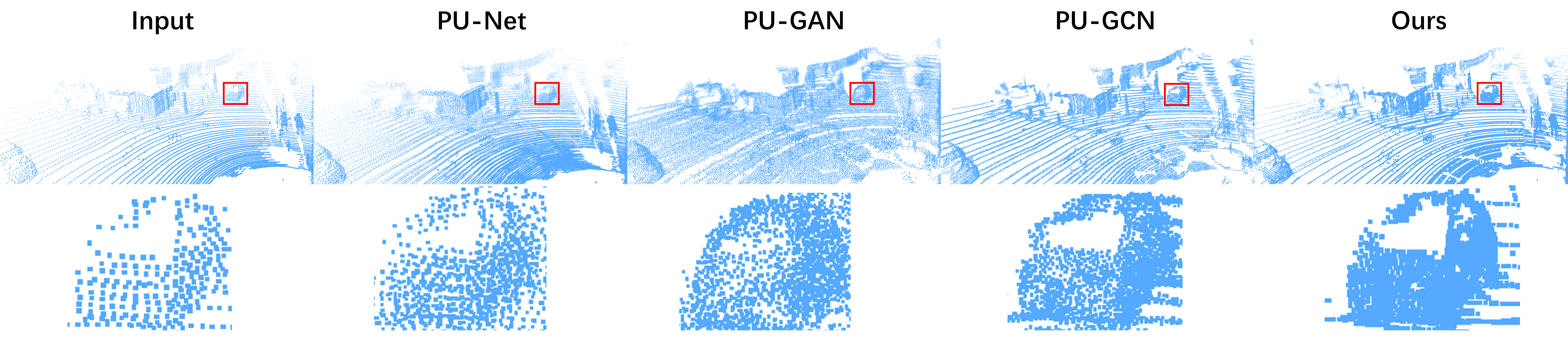}
    \vspace{-0.8cm}
    \caption{$4\times$ result on KITTI dataset. Compared with others, our method has less noise and more precisely reconstructs people and vehicle.}
    \label{fig:4X-KITTI}
\vspace{-0.3cm}
\end{figure*}
\begin{figure*}[t]
    \centering
    \includegraphics[width=\linewidth]{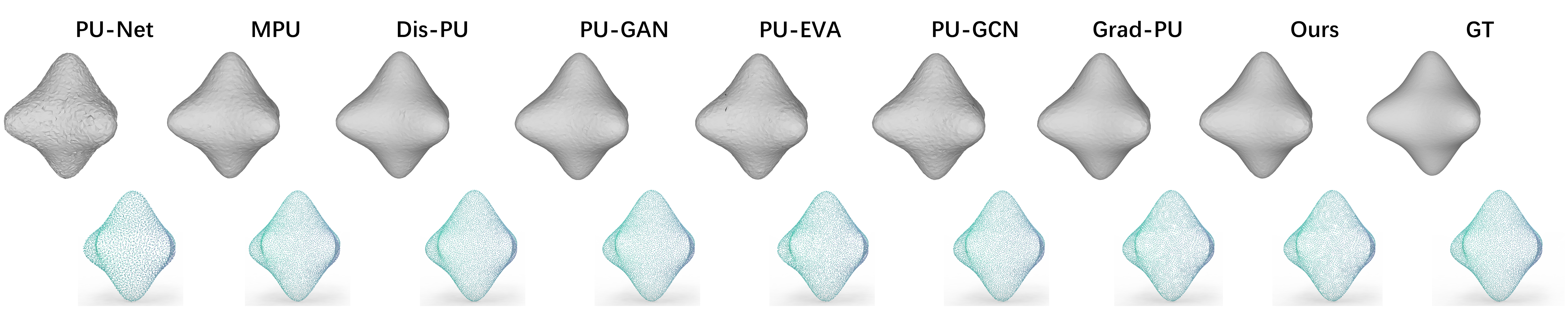}
    \vspace{-0.8cm}
    \caption{Meshes reconstructed by BallPivoting algorithm on PU-GAN dataset and their corresponding point clouds. Although all methods seem similar from the point clouds perspective, our method demonstrates better surface reconstruction capability with a smoother surface shown in the mesh. }
    \label{fig:surface reconstruction}
\vspace{-0.4cm}
\end{figure*}

\begin{table}[t]\footnotesize
\centering
\caption{Comparison between our method and others. Results are on the PU-GAN dataset with 4$\times$ upsampling rate.}
\vspace{-0.3cm}
{%
\begin{tabular}{llllllllll}
\toprule[1.5pt]
\multirow{3}{*}{Methods}       & \multicolumn{3}{c}{\multirow{3}{*}{\shortstack{CD$\downarrow$ \\ $10^{-3}$ }}} & \multicolumn{3}{c}{\multirow{3}{*}{\shortstack{F-score$\uparrow$\\@0.5\%}}} & \multicolumn{3}{c}{\multirow{3}{*}{\shortstack{P2F$\downarrow$ \\ $10^{-3}$ }}} \\ 
                               & \multicolumn{3}{c}{}                    & \multicolumn{3}{c}{}                    & \multicolumn{3}{c}{}                                     \\ & \multicolumn{3}{c}{}                    & \multicolumn{3}{c}{}                    & \multicolumn{3}{c}{}                                     \\ \midrule[1pt]
\multirow{2}{*}{PU-Net~\cite{yu2018pu}}        & \multicolumn{3}{l}{\multirow{2}{*}{0.4969}}   & \multicolumn{3}{l}{\multirow{2}{*}{0.0513}}   & \multicolumn{3}{l}{\multirow{2}{*}{8.8885}}  \\
                               & \multicolumn{3}{l}{}                    & \multicolumn{3}{l}{}                      & \multicolumn{3}{l}{}                                           \\
\multirow{2}{*}{MPU~\cite{yifan2019patch}}           & \multicolumn{3}{l}{\multirow{2}{*}{0.2725}}   &
\multicolumn{3}{l}{\multirow{2}{*}{0.1303}}     & \multicolumn{3}{l}{\multirow{2}{*}{3.1458}}       \\
                               & \multicolumn{3}{l}{}                    & \multicolumn{3}{l}{}                    & \multicolumn{3}{l}{}                                      \\
\multirow{2}{*}{PU-GAN~\cite{li2019pu}}        & \multicolumn{3}{l}{\multirow{2}{*}{0.2816}}   & \multicolumn{3}{l}{\multirow{2}{*}{0.1631}}   & \multicolumn{3}{l}{\multirow{2}{*}{1.9705}}     \\
                               & \multicolumn{3}{l}{}                    & \multicolumn{3}{l}{}                      & \multicolumn{3}{l}{}                                           \\
\multirow{2}{*}{Dis-PU~\cite{li2021point}}        & \multicolumn{3}{l}{\multirow{2}{*}{0.2716}}   & \multicolumn{3}{l}{\multirow{2}{*}{0.1428}}   & \multicolumn{3}{l}{\multirow{2}{*}{2.0247}}     \\
                               & \multicolumn{3}{l}{}                    & \multicolumn{3}{l}{}                      & \multicolumn{3}{l}{}                                        \\
\multirow{2}{*}{PU-EVA~\cite{luo2021pu}}        & \multicolumn{3}{l}{\multirow{2}{*}{0.3443}}   & \multicolumn{3}{l}{\multirow{2}{*}{0.1258}}   & \multicolumn{3}{l}{\multirow{2}{*}{3.4077}}     \\
                               & \multicolumn{3}{l}{}                    & \multicolumn{3}{l}{}                    & \multicolumn{3}{l}{}                                         \\
\multirow{2}{*}{PU-GCN~\cite{qian2021pu}}        & \multicolumn{3}{l}{\multirow{2}{*}{0.2712}}   & \multicolumn{3}{l}{\multirow{2}{*}{0.1385}}   & \multicolumn{3}{l}{\multirow{2}{*}{2.7706}}     \\
                               & \multicolumn{3}{l}{}                    & \multicolumn{3}{l}{}                      & \multicolumn{3}{l}{}                                         \\
\multirow{2}{*}{Grad-PU~\cite{he2023grad}}       & \multicolumn{3}{l}{\multirow{2}{*}{0.2657}}   & \multicolumn{3}{l}{\multirow{2}{*}{0.1636}}   & \multicolumn{3}{l}{\multirow{2}{*}{1.9250}}      \\
                               & \multicolumn{3}{l}{}                    & \multicolumn{3}{l}{}                    & \multicolumn{3}{l}{}                                       \\ \midrule[1pt]
\multirow{2}{*}{\textbf{Ours}} & \multicolumn{3}{l}{\multirow{2}{*}{\textbf{0.2653}}}   & \multicolumn{3}{l}{\multirow{2}{*}{\textbf{0.1638}}}   & \multicolumn{3}{l}{\multirow{2}{*}{\textbf{1.9105}}}   \\
                               & \multicolumn{3}{l}{}                    & \multicolumn{3}{l}{}                    & \multicolumn{3}{l}{}                                          \\ \bottomrule[1.5pt]
\end{tabular}}
\label{tab:compare}
\vspace{-0.6cm}
\end{table}

\begin{table}[t]
\centering
\caption{Performance of Grad-PU and ours on the PU-GAN dataset under different upsampling rates.}
\vspace{-0.3cm}
\resizebox{\linewidth}{!}{
\begin{tabular}{clllllllll|lllllllll}
\toprule[1.5pt]
\multirow{2}{*}{Methods}       & \multicolumn{9}{c|}{\multirow{2}{*}{Grad-PU~\cite{he2023grad}}}                                                                                 & \multicolumn{9}{c}{\multirow{2}{*}{Ours}}                                                                                    \\
                               & \multicolumn{9}{c|}{}                                                                                                         & \multicolumn{9}{c}{}                                                                                                         \\ \cline{2-19} 
\multirow{3}{*}{Rates}         & \multicolumn{3}{c}{\multirow{3}{*}{\shortstack{CD$\downarrow$ \\ $10^{-3}$ }}} & \multicolumn{3}{c}{\multirow{3}{*}{\shortstack{F-score$\uparrow$\\@0.5\%}}} & \multicolumn{3}{c|}{\multirow{3}{*}{\shortstack{P2F$\downarrow$ \\ $10^{-3}$ }}} & \multicolumn{3}{c}{\multirow{3}{*}{\shortstack{CD$\downarrow$ \\ $10^{-3}$ }}} & \multicolumn{3}{c}{\multirow{3}{*}{\shortstack{F-score$\uparrow$\\@0.5\%}}} & \multicolumn{3}{c}{\multirow{3}{*}{\shortstack{P2F$\downarrow$ \\ $10^{-3}$ }}} \\
                               & \multicolumn{3}{c}{}                    & \multicolumn{3}{c}{}                    & \multicolumn{3}{c|}{}                     & \multicolumn{3}{c}{}                    & \multicolumn{3}{c}{}                    & \multicolumn{3}{c}{}                     \\  & \multicolumn{3}{c}{}                    & \multicolumn{3}{c}{}                    & \multicolumn{3}{c|}{}                     & \multicolumn{3}{c}{}                    & \multicolumn{3}{c}{}                    & \multicolumn{3}{c}{}                     \\ \midrule[1pt]
\multirow{2}{*}{2$\times$}             & \multicolumn{3}{l}{\multirow{2}{*}{0.5362}}   & \multicolumn{3}{l}{\multirow{2}{*}{0.0881}}   & \multicolumn{3}{l|}{\multirow{2}{*}{1.1172}}    & \multicolumn{3}{l}{\multirow{2}{*}{\textbf{0.5357}}}   & \multicolumn{3}{l}{\multirow{2}{*}{\textbf{0.0884}}}   & \multicolumn{3}{l}{\multirow{2}{*}{\textbf{1.0325}}}    \\
                               & \multicolumn{3}{l}{}                    & \multicolumn{3}{l}{}                    & \multicolumn{3}{l|}{}                     & \multicolumn{3}{l}{}                    & \multicolumn{3}{l}{}                    & \multicolumn{3}{l}{}                     \\
\multirow{2}{*}{3$\times$}             & \multicolumn{3}{l}{\multirow{2}{*}{0.3447}}   & \multicolumn{3}{l}{\multirow{2}{*}{0.1362}}   & \multicolumn{3}{l|}{\multirow{2}{*}{1.4689}}    & \multicolumn{3}{l}{\multirow{2}{*}{\textbf{0.3446}}}   & \multicolumn{3}{l}{\multirow{2}{*}{\textbf{0.1367}}}   & \multicolumn{3}{l}{\multirow{2}{*}{\textbf{1.4382}}}    \\
                               & \multicolumn{3}{l}{}                    & \multicolumn{3}{l}{}                    & \multicolumn{3}{l|}{}                     & \multicolumn{3}{l}{}                    & \multicolumn{3}{l}{}                    & \multicolumn{3}{l}{}                     \\
\multirow{2}{*}{5$\times$}             & \multicolumn{3}{l}{\multirow{2}{*}{0.2406}}   & \multicolumn{3}{l}{\multirow{2}{*}{0.1972}}   & \multicolumn{3}{l|}{\multirow{2}{*}{2.5582}}    & \multicolumn{3}{l}{\multirow{2}{*}{\textbf{0.2403}}}   & \multicolumn{3}{l}{\multirow{2}{*}{\textbf{0.1973}}}   & \multicolumn{3}{l}{\multirow{2}{*}{\textbf{2.5474}}}    \\
                               & \multicolumn{3}{l}{}                    & \multicolumn{3}{l}{}                    & \multicolumn{3}{l|}{}                     & \multicolumn{3}{l}{}                    & \multicolumn{3}{l}{}                    & \multicolumn{3}{l}{}                     \\
\multirow{2}{*}{6$\times$}             & \multicolumn{3}{l}{\multirow{2}{*}{0.2303}}   & \multicolumn{3}{l}{\multirow{2}{*}{0.2136}}   & \multicolumn{3}{l|}{\multirow{2}{*}{3.0392}}    & \multicolumn{3}{l}{\multirow{2}{*}{\textbf{0.2301}}}   & \multicolumn{3}{l}{\multirow{2}{*}{\textbf{0.2138}}}   & \multicolumn{3}{l}{\multirow{2}{*}{\textbf{3.0280}}}    \\
                               & \multicolumn{3}{l}{}                    & \multicolumn{3}{l}{}                    & \multicolumn{3}{l|}{}                     & \multicolumn{3}{l}{}                    & \multicolumn{3}{l}{}                    & \multicolumn{3}{l}{}                     \\
\multirow{2}{*}{7$\times$}             & \multicolumn{3}{l}{\multirow{2}{*}{0.2264}}   & \multicolumn{3}{l}{\multirow{2}{*}{0.2376}}   & \multicolumn{3}{l|}{\multirow{2}{*}{3.4609}}    & \multicolumn{3}{l}{\multirow{2}{*}{\textbf{0.2261}}}   & \multicolumn{3}{l}{\multirow{2}{*}{\textbf{0.2378}}}   & \multicolumn{3}{l}{\multirow{2}{*}{\textbf{3.4500}}}    \\
                               & \multicolumn{3}{l}{}                    & \multicolumn{3}{l}{}                    & \multicolumn{3}{l|}{}                     & \multicolumn{3}{l}{}                    & \multicolumn{3}{l}{}                    & \multicolumn{3}{l}{}                     \\
\multirow{2}{*}{8$\times$}             & \multicolumn{3}{l}{\multirow{2}{*}{0.2262}}   & \multicolumn{3}{l}{\multirow{2}{*}{0.2512}}   & \multicolumn{3}{l|}{\multirow{2}{*}{3.8730}}    & \multicolumn{3}{l}{\multirow{2}{*}{\textbf{0.2259}}}   & \multicolumn{3}{l}{\multirow{2}{*}{\textbf{0.2514}}}   & \multicolumn{3}{l}{\multirow{2}{*}{\textbf{3.8629}}}    \\
                               & \multicolumn{3}{l}{}                    & \multicolumn{3}{l}{}                    & \multicolumn{3}{l|}{}                     & \multicolumn{3}{l}{}                    & \multicolumn{3}{l}{}                    & \multicolumn{3}{l}{}                     \\ \bottomrule[1.5pt]
\end{tabular}
}
\vspace{-0.3cm}
\label{tab:rates}
\end{table}

\subsection{Loss function}
The overall loss function $\boldsymbol{\mathit{L}}$ comprises three components.
The first part is the L1 loss $\boldsymbol{\mathit{L_{d}}}$ between the predicted points obtained from the MBPU network and the ground truth. 
The second part is the view loss $\boldsymbol{\mathit{L_{v}}}$ obtained through point cloud rendering. 
The third part is the Chamfer Distance loss $\boldsymbol{\mathit{L_{cd}}}$ between the predicted points and the ground truth $\boldsymbol{\mathit{Q}}$.
The specific formulas are as follows:
\begin{equation}\small
\label{eq11}
\boldsymbol{\mathit{L=L_{d}+\alpha L_{v}+\beta L_{cd} }},
\end{equation}
\begin{equation}\small
\label{eq12}
\boldsymbol{\mathit{L_{d}(P_{I},Q)=\frac{1}{|P_{I}|}\sum_{p\in P_{I},q\in Q}\|q-\text{MBPU}(p)\|}},
\end{equation}
\begin{align}\small
\label{eq13}
\boldsymbol{\mathit{L_{CD}(P_{I}, Q)}} &= \frac{1}{\boldsymbol{\mathit{|P_{I}|}}} \sum_{p \in \boldsymbol{\mathit{P_{I}}}} \min_{q \in \boldsymbol{\mathit{Q}}} \| \boldsymbol{\mathit{p}} - \boldsymbol{\mathit{q}} \|_2^2 \nonumber \\
&\quad + \frac{1}{\boldsymbol{\mathit{|Q|}}} \sum_{q \in \boldsymbol{\mathit{Q}}} \min_{p \in \boldsymbol{\mathit{P_{I}}}} \| \boldsymbol{\mathit{q}} - \boldsymbol{\mathit{p}} \|_2^2.
\end{align}

\begin{table}[t]
\centering
\caption{4$\times$ quantitative results on PU1K dataset with different methods.}
\vspace{-0.3cm}
\begin{tabular}{clllllllllllllllllllllll}
\toprule[1.5pt]
\multicolumn{6}{l}{\multirow{2}{*}{Methods}} & \multicolumn{6}{c}{\multirow{2}{*}{\shortstack{CD$\downarrow$ \\ $10^{-3}$ }}} & \multicolumn{6}{c}{\multirow{2}{*}{\shortstack{HD$\downarrow$ \\ $10^{-3}$ }}} & \multicolumn{6}{c}{\multirow{2}{*}{\shortstack{P2F$\downarrow$ \\ $10^{-3}$ }}} \\
\multicolumn{6}{c}{}                         & \multicolumn{6}{c}{}                    & \multicolumn{6}{c}{}                    & \multicolumn{6}{c}{}                     \\ \midrule[1pt]
\multicolumn{6}{l}{PU-Net~\cite{yu2018pu}}                   & \multicolumn{6}{l}{1.155}                    & \multicolumn{6}{l}{15.170}                    & \multicolumn{6}{l}{4.834}                     \\
\multicolumn{6}{l}{MPU~\cite{yifan2019patch}}                      & \multicolumn{6}{l}{0.935}                    & \multicolumn{6}{l}{13.327}                    & \multicolumn{6}{l}{3.511}                     \\
\multicolumn{6}{l}{PU-GCN~\cite{qian2021pu}}                   & \multicolumn{6}{l}{0.585}                    & \multicolumn{6}{l}{7.577}                    & \multicolumn{6}{l}{2.499}                     \\
\multicolumn{6}{l}{Grad-PU~\cite{he2023grad}}                  & \multicolumn{6}{l}{0.422}                    & \multicolumn{6}{l}{4.307}                    & \multicolumn{6}{l}{1.561}                     \\ \midrule[1pt]
\multicolumn{6}{l}{\textbf{Ours}}            & \multicolumn{6}{l}{\textbf{0.413}}                    & \multicolumn{6}{l}{\textbf{4.019}}                    & \multicolumn{6}{l}{\textbf{1.479}}                     \\ \bottomrule[1.5pt]
\end{tabular}
\label{tab:PU1K}
\vspace{-0.5cm}
\end{table}

\section{Experiments}

\subsection{Comparison with SOTA}
\subsubsection{\textbf{Evaluation on PU-GAN Dataset}}
We evaluate the performance of our model against existing methods using the PU-GAN dataset at a 4X upsampling rate.
All the comparative models were trained using their official codes and default settings for fairness.
Table~\ref{tab:compare} demonstrates that our method surpasses other models in terms of CD~\cite{thayananthan2003shape}, F-score~\cite{knapitsch2017tanks}, and P2F~\cite{achlioptas2018learning} metrics.
The visualization results of the PU-GAN experiment are shown in Fig~\ref{fig:4X-PUGAN}.
In the case of the finger surface, our method yields results that closely resemble the ground truth, while other methods tend to exhibit some outliers.
Both the quantitative results and the visualizations demonstrate that our method outperforms previous approaches.

\subsubsection{\textbf{Evaluation with Arbitrary Upsampling Rates}}
By harnessing the flexibility of midpoint interpolation in terms of the number of points, our model can perform upsampling at varying rates within a single training session.
Similarly, the prior method Grad-PU also features arbitrary upsampling rates.
So we compare the Grad-PU across different upsampling rates.
It is important to mention that the model trained for this experiment is identical to the one utilized in the PUGAN experiment.
During testing, except for the upsampling rate, all parameters remain unchanged.
Furthermore, all parameters of both Grad-PU and our method are kept consistent.
As demonstrated in Table~\ref{tab:rates}, our model surpasses Grad-PU across all upsampling rates ranging from 2 to 8.

\subsubsection{\textbf{Evaluation on PU1K Dataset}}
We also carry out a 4X upsampling evaluation on the PU1K dataset.
The results of PU-Net, MPU, and PU-GCN are directly extracted from their original papers, whereas Grad-PU and our method are trained using the same parameters with 10 epochs.
As presented in Table~\ref{tab:PU1K}, our model surpasses other methods across all metrics.
The visualization results are depicted in Fig~\ref{fig:4X-PU1K}, where it can be observed that other methods produce outliers and artifacts around the surface.
In contrast, our method reconstructs objects' surfaces more accurately with fewer outliers.
\par Moreover, in this experiment, our model demonstrates rapid convergence speed. 
Its convergence in less than 10 epochs is attributed to the strong representative learning and fast training advantages of the Mamba block.

\subsubsection{\textbf{Evaluation on KITTI~\cite{geiger2013vision} Dataset}}
To validate our model in zero-shot and real-world settings, we also assess its performance on a real-world dataset using the model trained on the PU-GAN dataset.
We conducted a visual comparison due to the absence of the ground truth point cloud.
Result is shown in Fig~\ref{fig:4X-KITTI}.
Our model showcases a cleaner point cloud that closely aligns with real-world scenarios.
Both the pedestrians and vehicles in the scenes are well-reconstructed.
In contrast, other methods generate upsampled point clouds with more noise and numerous artifacts.

\subsection{Surface Reconstruction}
To vaildate our model's performance on downstream tasks, we apply surface reconstruction to the PU-GAN dataset.
We utilize BallPivoting~\cite{bernardini1999ball} algorithm to reconstruct meshes on the upsampled point clouds from the 4X PU-GAN experiment.
Although the variances between methods are subtle at the point cloud level in this instance, the mesh visualization in Fig.~\ref{fig:surface reconstruction} distinctly illustrates that our method surpasses the others, yielding smoother surfaces that closely match the actual geometry.

\subsection{Robustness Test}
Given that point clouds captured in real-world scenarios often contain noise, we evaluate the model's robustness by introducing noise into the dataset.
The robustness test is performed on the PU-GAN dataset with a 4x upsampling rate.
The process of adding noise to the dataset is following~\cite{he2023grad}.
Specifically, The noises are generated by sampling numbers from a standard Gaussian distribution $\mathcal{N}(0, 1)$ and then multiplying them by a noise level $\tau$.evel $\tau$.  
The results are presented in Table \ref{tab:noisy}. Our approach demonstrates superior performance compared to other methods across all metrics at noise levels of 0.005 and 0.01.

\section{Ablation Study}

\begin{table}[t]
\centering
\caption{$4\times$ comparative results on PU-GAN dataset with a different loss function.}
\vspace{-0.3cm}
\begin{tabular}{cccccc|cccccc}
\toprule[1.5pt]
\multicolumn{2}{c}{\multirow{2}{*}{loss\_p2p}} & \multicolumn{2}{c}{\multirow{2}{*}{loss\_cd}} & \multicolumn{2}{c|}{\multirow{2}{*}{loss\_view}} & \multicolumn{2}{c}{\multirow{2}{*}{\shortstack{CD$\downarrow$ \\ $10^{-3}$ }}} & \multicolumn{2}{c}{\multirow{2}{*}{\shortstack{F-score$\uparrow$\\@1\%}}} & \multicolumn{2}{c}{\multirow{2}{*}{\shortstack{P2F$\downarrow$ \\ $10^{-3}$ }}} \\
\multicolumn{2}{c}{}                           & \multicolumn{2}{c}{}                          & \multicolumn{2}{c|}{}                            & \multicolumn{2}{c}{}                    & \multicolumn{2}{c}{}                        & \multicolumn{2}{c}{}                     \\ \midrule[1pt]
\multicolumn{2}{c}{\multirow{2}{*}{\checkmark}}          & \multicolumn{2}{c}{\multirow{2}{*}{}}         & \multicolumn{2}{c|}{\multirow{2}{*}{\checkmark}}           & \multicolumn{2}{c}{\multirow{2}{*}{0.2655}}   & \multicolumn{2}{c}{\multirow{2}{*}{0.5881}}       & \multicolumn{2}{c}{\multirow{2}{*}{1.9110}}    \\
\multicolumn{2}{c}{}                           & \multicolumn{2}{c}{}                          & \multicolumn{2}{c|}{}                            & \multicolumn{2}{c}{}                    & \multicolumn{2}{c}{}                        & \multicolumn{2}{c}{}                     \\
\multicolumn{2}{c}{\multirow{2}{*}{\checkmark}}          & \multicolumn{2}{c}{\multirow{2}{*}{\checkmark}}         & \multicolumn{2}{c|}{\multirow{2}{*}{}}           & \multicolumn{2}{c}{\multirow{2}{*}{0.2656}}   & \multicolumn{2}{c}{\multirow{2}{*}{0.5881}}       & \multicolumn{2}{c}{\multirow{2}{*}{\textbf{1.9040}}}    \\
\multicolumn{2}{c}{}                           & \multicolumn{2}{c}{}                          & \multicolumn{2}{c|}{}                            & \multicolumn{2}{c}{}                    & \multicolumn{2}{c}{}                        & \multicolumn{2}{c}{}                     \\
\multicolumn{2}{c}{\multirow{2}{*}{\checkmark}}          & \multicolumn{2}{c}{\multirow{2}{*}{\checkmark}}         & \multicolumn{2}{c|}{\multirow{2}{*}{\checkmark}}           & \multicolumn{2}{c}{\multirow{2}{*}{\textbf{0.2653}}}   & \multicolumn{2}{c}{\multirow{2}{*}{\textbf{0.5882}}}       & \multicolumn{2}{c}{\multirow{2}{*}{1.9105}}    \\
\multicolumn{2}{c}{}                           & \multicolumn{2}{c}{}                          & \multicolumn{2}{c|}{}                            & \multicolumn{2}{c}{}                    & \multicolumn{2}{c}{}                        & \multicolumn{2}{c}{}                     \\ \bottomrule[1.5pt]
\end{tabular}
\label{tab:ablation}
\vspace{-0.5cm}
\end{table}

\begin{table}[t]
\caption{$4\times$ results on PU-GAN dataset with different noise level $\tau$.}
\vspace{-0.3cm}
\resizebox{\linewidth}{!}{
\begin{tabular}{lllllllllllllll}
\toprule[1.5pt]
\multicolumn{3}{l}{\multirow{2}{*}{Noise Levels}}  & \multicolumn{6}{c}{\multirow{2}{*}{$\tau = 0.005$}}                                                                                                                                   & \multicolumn{6}{c}{\multirow{2}{*}{$\tau = 0.01$}}                                                                   \\
\multicolumn{3}{l}{}                               & \multicolumn{6}{c}{}                                                                                                                                                                               & \multicolumn{6}{c}{}                                                                                                              \\ \cline{4-15} 
\multicolumn{3}{l}{\multirow{3}{*}{Methods}}       & \multicolumn{2}{c}{\multirow{3}{*}{\shortstack{CD$\downarrow$ \\ $10^{-3}$ }}} & \multicolumn{2}{c}{\multirow{3}{*}{\shortstack{F-score$\uparrow$\\@1\%}}} & \multicolumn{2}{c}{\multirow{3}{*}{\shortstack{P2F$\downarrow$ \\ $10^{-3}$}}} & \multicolumn{2}{c}{\multirow{3}{*}{\shortstack{CD$\downarrow$ \\ $10^{-3}$}}} & \multicolumn{2}{c}{\multirow{3}{*}{\shortstack{F-score$\uparrow$\\@1\%}}} & \multicolumn{2}{c}{\multirow{3}{*}{\shortstack{P2F$\downarrow$ \\ $10^{-3}$}}} \\
\multicolumn{3}{l}{}                               & \multicolumn{2}{c}{}                                                                                     & \multicolumn{2}{c}{}                         & \multicolumn{2}{c}{}                     & \multicolumn{2}{c}{}                    & \multicolumn{2}{c}{}                         & \multicolumn{2}{c}{}                     \\
\multicolumn{3}{l}{}                               & \multicolumn{2}{c}{}                                                                                     & \multicolumn{2}{c}{}                         & \multicolumn{2}{c}{}                     & \multicolumn{2}{c}{}                    & \multicolumn{2}{c}{}                         & \multicolumn{2}{c}{}                     \\ \midrule[1pt]
\multicolumn{3}{l}{\multirow{2}{*}{PU-Net~\cite{yu2018pu}}}        & \multicolumn{2}{c}{\multirow{2}{*}{0.5241}}                                                                    & \multicolumn{2}{c}{\multirow{2}{*}{0.2962}}        & \multicolumn{2}{c}{\multirow{2}{*}{9.4178}}    & \multicolumn{2}{c}{\multirow{2}{*}{0.5807}}   & \multicolumn{2}{c}{\multirow{2}{*}{0.2659}}        & \multicolumn{2}{c}{\multirow{2}{*}{10.5629}}    \\
\multicolumn{3}{l}{}                               & \multicolumn{2}{l}{}                                                                                     & \multicolumn{2}{l}{}                         & \multicolumn{2}{l}{}                     & \multicolumn{2}{l}{}                    & \multicolumn{2}{l}{}                         & \multicolumn{2}{l}{}                     \\
\multicolumn{3}{l}{\multirow{2}{*}{MPU~\cite{yifan2019patch}}}           & \multicolumn{2}{c}{\multirow{2}{*}{0.4314}}                                                                    & \multicolumn{2}{c}{\multirow{2}{*}{0.3661}}        & \multicolumn{2}{c}{\multirow{2}{*}{7.1216}}    & \multicolumn{2}{c}{\multirow{2}{*}{0.5526}}   & \multicolumn{2}{c}{\multirow{2}{*}{0.3005}}        & \multicolumn{2}{c}{\multirow{2}{*}{9.1171}}    \\
\multicolumn{3}{l}{}                               & \multicolumn{2}{l}{}                                                                                     & \multicolumn{2}{l}{}                         & \multicolumn{2}{l}{}                     & \multicolumn{2}{l}{}                    & \multicolumn{2}{l}{}                         & \multicolumn{2}{l}{}                     \\
\multicolumn{3}{l}{\multirow{2}{*}{PU-GAN~\cite{li2019pu}}}        & \multicolumn{2}{c}{\multirow{2}{*}{0.5455}}                                                                    & \multicolumn{2}{c}{\multirow{2}{*}{0.2883}}        & \multicolumn{2}{c}{\multirow{2}{*}{10.0741}}    & \multicolumn{2}{c}{\multirow{2}{*}{0.7231}}   & \multicolumn{2}{c}{\multirow{2}{*}{0.2358}}        & \multicolumn{2}{c}{\multirow{2}{*}{12.5321}}    \\
\multicolumn{3}{l}{}                               & \multicolumn{2}{l}{}                                                                                     & \multicolumn{2}{l}{}                         & \multicolumn{2}{l}{}                     & \multicolumn{2}{l}{}                    & \multicolumn{2}{l}{}                         & \multicolumn{2}{l}{}                     \\
\multicolumn{3}{l}{\multirow{2}{*}{PU-GCN~\cite{qian2021pu}}}        & \multicolumn{2}{c}{\multirow{2}{*}{0.4355}}                                                                    & \multicolumn{2}{c}{\multirow{2}{*}{0.3710}}        & \multicolumn{2}{c}{\multirow{2}{*}{7.0081}}    & \multicolumn{2}{c}{\multirow{2}{*}{0.5479}}   & \multicolumn{2}{c}{\multirow{2}{*}{0.3024}}        & \multicolumn{2}{c}{\multirow{2}{*}{9.0406}}    \\
\multicolumn{3}{l}{}                               & \multicolumn{2}{l}{}                                                                                     & \multicolumn{2}{l}{}                         & \multicolumn{2}{l}{}                     & \multicolumn{2}{l}{}                    & \multicolumn{2}{l}{}                         & \multicolumn{2}{l}{}                     \\
\multicolumn{3}{l}{\multirow{2}{*}{Grad-PU~\cite{he2023grad}}}       & \multicolumn{2}{c}{\multirow{2}{*}{0.3076}}                                                                    & \multicolumn{2}{c}{\multirow{2}{*}{0.5138}}        & \multicolumn{2}{c}{\multirow{2}{*}{4.0931}}    & \multicolumn{2}{c}{\multirow{2}{*}{0.4363}}   & \multicolumn{2}{c}{\multirow{2}{*}{0.3760}}        & \multicolumn{2}{c}{\multirow{2}{*}{6.8198}}    \\
\multicolumn{3}{l}{}                               & \multicolumn{2}{l}{}                                                                                     & \multicolumn{2}{l}{}                         & \multicolumn{2}{l}{}                     & \multicolumn{2}{l}{}                    & \multicolumn{2}{l}{}                         & \multicolumn{2}{l}{}                     \\ \midrule[1pt]
\multicolumn{3}{l}{\multirow{2}{*}{\textbf{Ours}}} & \multicolumn{2}{c}{\multirow{2}{*}{\textbf{0.3074}}}                                                                    & \multicolumn{2}{c}{\multirow{2}{*}{\textbf{0.5143}}}        & \multicolumn{2}{c}{\multirow{2}{*}{\textbf{4.0794}}}    & \multicolumn{2}{c}{\multirow{2}{*}{\textbf{0.4356}}}   & \multicolumn{2}{c}{\multirow{2}{*}{\textbf{0.3765}}}        & \multicolumn{2}{c}{\multirow{2}{*}{\textbf{6.7988}}}    \\
\multicolumn{3}{l}{}                               & \multicolumn{2}{l}{}                                                                                     & \multicolumn{2}{l}{}                         & \multicolumn{2}{l}{}                     & \multicolumn{2}{l}{}                    & \multicolumn{2}{l}{}                         & \multicolumn{2}{l}{}                     \\ \bottomrule[1.5pt]
\end{tabular}
}
\label{tab:noisy}
\vspace{-0.6cm}
\end{table}

In this section, we perform ablation studies on the PU-GAN dataset with a 4X upsampling rate.

\subsection{Impact of Differentiable Rendering: With vs. Without}
We evaluate the influence of the differentiable render module by comparing model metrics with and without its incorporation.
The results are presented in the first and third rows of Table~\ref{tab:ablation}.
Our model exhibits superior performance according to the CD and F-score metrics, albeit showing slightly poorer results on P2F.
Given that P2F is asymmetrical, covering only the transition from upsampled points to the ground truth surfaces and not the reverse direction, as highlighted in previous studies~\cite{li2019pu,he2023grad}, we can still conclude that our model, incorporating differentiable rendering, outperforms in terms of CD and F-score metrics.
\vspace{-0.3cm}
\subsection{Impact of Chamfer Distance Loss: With vs. Without}
We also evaluate the influence of the chamfer distance loss in the loss function.
Table~\ref{tab:ablation} illustrates that our model incorporating the chamfer distance loss outperforms the model without its inclusion across all metrics. 
This indicates that the constraints imposed by the chamfer distance loss could help regress the 3D point shift, thereby enhancing our model's ability to refine the point cloud.

\section{Conclusion}
In this paper, we introduce MBPU, the first plug-and-play state space model designed for point cloud upsampling.
We utilize midpoint interpolation to achieve flexible upsampling rates during a single training process. 
We design the MBPU network to enhance point position refinement, utilizing the Mamba block as the backbone and benefiting from the state space model to achieve excellent performance in long sequences and rapid convergence speed.
Moreover, we introduce two branches - point-to-point distance and 3D position shift - for prediction in gradient descent, effectively preserving local details while upholding the global structure.
We also introduce a fast differentiable renderer to refine the 3D position shift, consequently reducing artifacts and outliers in the output.
Extensive experiments have validated that our method surpasses previous state-of-the-art techniques, attaining superior metrics and visual outcomes.





\bibliographystyle{ieeetr}
\bibliography{ref}

\end{document}